# Learning to Mine Chinese Coordinate Terms Using the Web

Xiaojiang Huang      Xiaojun Wan      Jianguo Xiao

*Institute of Computer Science & Technology*

*The MOE Key Laboratory of Computational Linguistics*

*Peking University, Beijing, China*

{huangxiaojiang, wanxiaojun, xiaojianguo}@pku.edu.cn

**Abstract.** Coordinate relation refers to the relation between instances of a concept and the relation between the directly hyponyms of a concept. In this paper, we focus on the task of extracting terms which are coordinate with a user given seed term in Chinese, and grouping the terms which belong to different concepts if the seed term has several meanings. We propose a semi-supervised method that integrates manually defined linguistic patterns and automatically learned semi-structural patterns to extract coordinate terms in Chinese from web search results. In addition, terms are grouped into different concepts based on their co-occurring terms and contexts. We further calculate the saliency scores of extracted terms and rank them accordingly. Experimental results demonstrate that our proposed method generates results with high quality and wide coverage.

## 1. Introduction

Any term is related with many other terms. The relations between terms are very important, and many efforts have been made to acquire these relations. One kind of the those relations is coordination, which refers to the relation among the instances of the same concept, or the direct hyponyms of the same concept, e.g. {*China*, *British*, *USA*, …}. Collections of coordinate terms are useful in many applications. For example, they can be directly used as recommendations in online shops and search engines. When a user is interested in *iPad*, it can be very useful to recommend him/her with *Playbook* or *GalaxyTab*. The coordinate terms are also fundamental resources in many NLP and web mining tasks, e.g. named entity extraction, question answering, comparative analysis, etc. Taking the comparative text mining as an example, the task aims to highlight the commonalities and differences among comparable objects. The comparative text mining usually consist of three steps: comparable objects finding, relevant document retrieval, and comparison results mining. The coordinate term mining is useful in both the first and the third steps. The coordinate terms belong to the same concept, and share many aspects, and thus they are potentially comparable objects, e.g., "*Iraq War* vs. *Afghanistan War*". The coordinate terms can also be used as indications of comparative points, e.g. "*sunny - storm*" indicates a comparison on the weather condition.

The Coordinate Term Mining task aims to extract the coordinate relations among terms. It has attracted much attention recently, and a few unsupervised and semi-supervised approaches have been proposed. The unsupervised methods use extraction patterns to find new elements of a given class, or use corpus-based term similarity to find term clusters. The semi-supervised methods start with a set of seed terms, and use patterns or distributional similarity to find terms similar to the seed terms. Some mining methods run offline. These systems analyze a corpus, extract all the coordinate terms, store them in databases or index structures, and then retrieve them from the database when needed. Some other methods run on the fly. When a user submits a query, the systems find a proper corpus and extract the coordinate terms instantly. Both service modes have advantages and drawbacks. The offline systems can access the whole corpus for many times, and thus machine learning techniques can be easily applied. However, due to time constraints of mass data processing, the offline methods mostly rely on general features for all the terms, and thus some useful query term dependent information may be ignored, which may lead to poor results for particular queries. In contrast, the online systems focus on the specified query term, and thus they can utilize the query-dependent information to help extract coordinate relations. In addition, they are able to deal with new queries and produce up-to-date results. The major disadvantages of online systems are as follows: 1) they can only access part of the corpus; 2) sophisticated techniques can barely be used due to the restriction of the response time. In this study, we focus on the on-the-fly mode.

Google Sets was once a well-known example of coordinate term search system[1]. As shown in fig. 1, it took a few coordinate terms (i.e. items from a set of things) as the query, and tried to predict more coordinate terms (i.e. other items in the set). Although it could run with only one query item, the results were actually quite messy. So far, most semi-supervised coordinate mining systems require several seed terms to start up, while those which can run with only one seed term often result in low coverage and/or accuracy. However, the ability to deal with only one seed term is useful in many fields. For example, it has been shown that the named entity recognition (NER) system can be benefit from coordinate terms of an entity (Wan et al. 2011). But in this case, the NER system is not able to provide two or more coordinate seed entities.

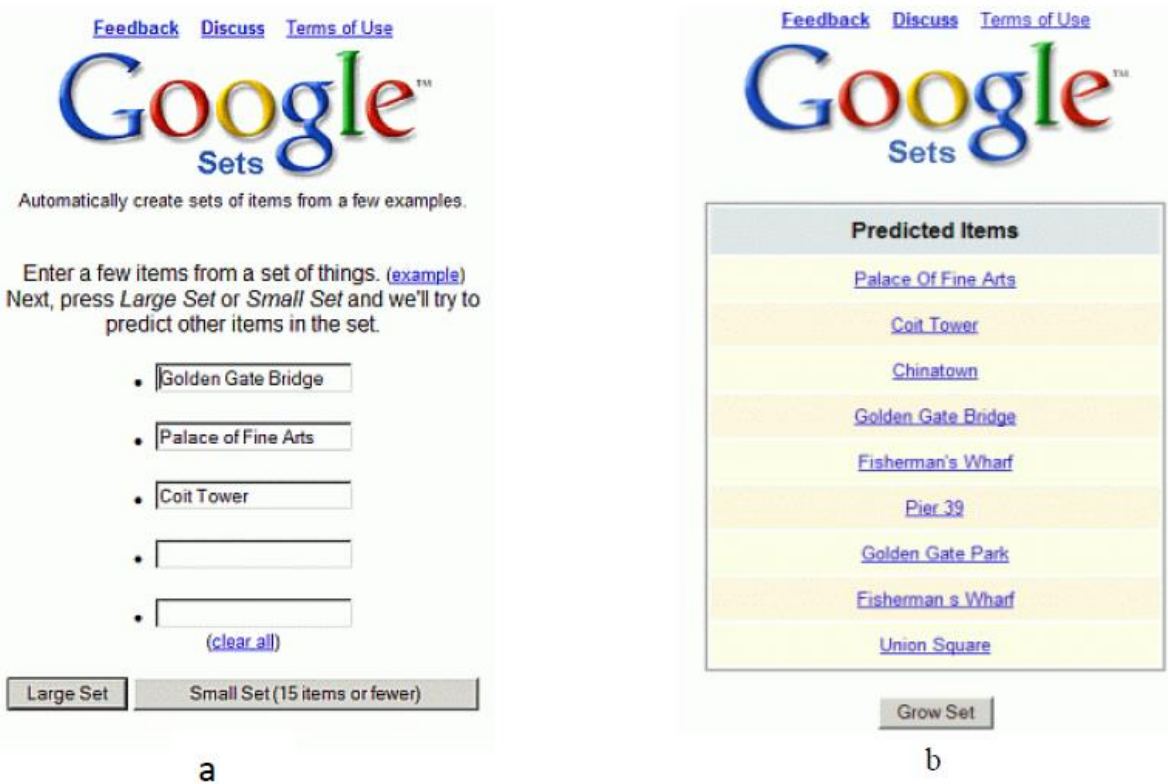

**Fig. 1**: a) The example inputs of Google Sets b) The example outputs of Google Sets

However, it is not easy to reduce the number of input seed terms to only one. Great challenges are brought by the minimization of the given query information. First, it is not easy to retrieve proper documents by using only one seed term. Without the clues of coordinate relations, the search engines usually return pages which do not contain any coordinate terms. Second, it is harder to extract coordinate terms from pages using one seed. Many existing information extraction algorithms need several different seeds to learn wrappers from their common contexts. Thus this kind of methods cannot be applied directly because there is no "common context" of a single seed. Third, and the most importantly, a term may be ambiguous, and thus the coordinate relationship is not clearly defined. For example, *Golden Gate Bridge* can refer to one of the tourist destinations in San Francisco. In this case, *Alcatraz*, *Union Square* and *Pier 39* are coordinate terms of *Golden Gate Bridge*. Meantime, *Golden Gate Bridge*" can also refer to one of the famous bridges in the world, and in such case, *Tower Bridge* and *Rialto* are possible coordinate terms. However, the coordinate terms in different cases are not coordinate with each other. Adding *Tower Bridge* into {*Alcatraz*, *Union Square*} will not be appropriate. In this case, a coordinate term mining system should find all possible results, and group them into several subsets of meanings accordingly.

In this study, we focus on the task of extracting coordinate terms of a Chinese seed term from the Web. The problem is formalized as the task of set expansion with only a single seed term. In this task, the seed term can be given in any open domain, e.g. an organization name, a location name, a product name, an event, etc. In our approach, we first submit several elaborate queries based on the seed term to a web search engine, and then exploit manually defined linguistic patterns and automatically learned semi-structural patterns to extract candidate terms from the search results. After that, we group terms into different concepts. Finally the terms are ranked according to their significance.

---

[1] It was shut down in September 2011.

As compared with previous works, the advantages of our approach are listed as follows:

1. The system needs only one seed term as the input, and thus it minimizes the requirement of prior knowledge from the user.
2. The system can achieve a high coverage of coordinate terms while keeping the high precision.
3. The system can automatically disambiguate different meanings in the terms.
4. The system requires no domain knowledge and returns up-to-date results, and thus it is suitable for open-domain information extraction applications.

The rest of this paper is organized as follows: In section 2 we give a brief review of related works. In section 3 we define the basic concepts and the mining task. The proposed coordinate term mining algorithm is described in section 4. In section 5 we evaluate our work in experiments. Section 6 introduces an application of our approach on named entity recognition in news comments. Finally, section 7 concludes the study and talks about the future works.

## 2. Related work

### 2.1. Term Extraction and Set Expansion

Extracting useful terms such as entity and feature names is a hot sub task of information extraction, and a number of methods have been proposed. Term extraction methods differ in degree of structure in the data, requirement of labeled examples, range of extraction targets and the types of features employed. Supervised approaches use large sets of labeled examples, perform focused extraction and employ sentence-level features. These methods are usually restricted in particular classes of terms (e.g. People, Organizations, and Locations) where large training sets are available (McCallum et al. 2000; McCallum and Li 2003). In our study, the user given seed term can be various, and it is too labor intensive to label examples manually. Unsupervised approaches need no labeled data but use extraction patterns to find new elements of a given class (Etzioni et al. 2005), or use corpus-based term similarity to find term clusters (Ghahramani and Heller 2006). These methods usually require large computing power and long time to deal with mass data, and thus they are more suitable for offline usage. Semi-supervised methods start with a set of seed terms, and use patterns or distributional similarity to find terms similar to the seed set (Jain and Pennacchiotti 2010; Ohshima et al. 2006; Pantel et al. 2009; Paşca 2007; Wang and Cohen 2007). The task of using a few seeds to find more terms is also called Set Expansion. These methods are useful for extending term classes where large labeled data sets are not available.

GoogleSets was once a well-known online system that does set expansion using Web search and mining techniques. The exact method underlying GoogleSets is unknown due to commercial privacy, but it most likely performs as described in (Tong and Dean 2008). The method first extracts a list by considering HTML tags, tables, commas or semicolons in web pages, and then ranks all items using the on-topic and off-topic models. SEAL is another online demo system for set expansion of named entities (Wang and Cohen 2007). Given a small number of seed objects, SEAL first downloads the web pages that containing the seed objects, and then automatically constructs wrappers for each page. New entities can be extracted by applying the wrappers on the web pages. The weakness of SEAL lies in that it requires at least two seed objects as input. Iterative SEAL has been proposed to allow a user to provide a large number of seeds (e.g. ten seeds) by making several calls to SEAL. We can see that neither SEAL nor Iterative SEAL can address the task of finding coordinate terms for a single term. Riloff and Shepherd (1997) propose to extract category members according to their occurrence within a small context window. This method needs a large corpus for counting occurrence relations. Though the model may deal with only one seed, it actually uses five seeds in the experiment. Besides, the authors claim that additional seed words tend to improve performance. Pantel et al. (2009) use word distributional similarity calculated on the Web-Scale corpus for set expansion. Thelen and Riloff (2002) propose a bootstrapping method to learn semantic lexicons for multiple categories. It also needs a text corpus and several seed words for each category to learn extraction patterns.

Set expansion with only one seed has attracted researchers' attention, and a few algorithms have been proposed. Ohshima et al. (2006) propose a method for searching coordinate terms using a conventional Web search engine to do two searches where queries are generated by combining the user's query term with a conjunction "OR". Li et al. use several manually written linguistic patterns to get the pages which may contain information of competitors and extract candidates (Bao et al. 2008; Li et al. 2006). Liu et al. (2007) use generic situation keywords extracted from a given news story as query to a web search engine to find comparable cases, and then extract comparable entities of the main entity in the given news story based on the similarity of their contexts. The weakness of these methods lies in that they usually generate only a few results. Jain and Pantel (2009) study a similar task, but use a different kind of approach. They first learn domain-independent patterns for extracting comparable relations by using a bootstrapping method, and then build a database of comparable pairs. The expansion task is then achieved by a simple database query.

Set expansion techniques have been successfully used for many NLP tasks, including named entity extraction

(Pennacchiotti and Pantel 2009; Wan et al. 2011), word sense disambiguation (Mihalcea et al. 2004), sentiment analysis (McIntosh 2010; Velikovich et al. 2010), question answering (Gupta and Sarawagi 2009; Wang et al. 2008), comparative analysis (Bao et al. 2008; Li et al. 2006), etc.

*2.2. Term Relation Extraction*

The coordinate relation between two terms can be considered as a particular kind of term relation. To date, there exist a number of previous works for extracting terms with a certain type of relation.

The Hyponym/Hypernym relation has been widely examined in previous works. Shinzato and Torisawa (2004) propose an automatic method for acquiring hyponymy relations from HTML documents on the Web by using clues such as itemization or listing in HTML documents and statistical measures such as document frequencies and verb-noun co-occurrence. Hearst (1992) proposes a method for the automatic acquisition of the hyponymy lexical relation from text by identifying a set of lexico-syntactic patterns. Snow et al. (2005) present a new algorithm for automatically learning hypernym relations from text by using dependency path features extracted from parse trees. Caraballo (1999) builds the hypernym-labeled noun hierarchy of WordNet by clustering nouns into a hierarchy using data on conjunctions and appositives appearing in the Wall Street Journal. Manning (1993) presents a method for producing a dictionary of sub-categorization frames from unlabeled text corpora. Sanderson and Croft (1999) present a means of automatically deriving a hierarchical organization of concepts from a set of documents by using a type of co-occurrence known as subsumption. Glover et al. (2002) create a statistical model for inferring hierarchical term relationships about a topic, given only a small set of example web pages on the topic.

A few researches have focused on extracting terms that have the same hypernym or are semantically similar. Shinzato and Torisawa (2005) provide a simple method to extract only semantically coherent itemizations from HTML documents. Lund and Burgess (1996) propose to construct semantic spaces in which a word is presented by a vector of co-occurred words, and calculate the semantic similarity between any pair of words using the distances of two vectors. Lin (1998) defines a word similarity measure based on the distributional pattern of words, and constructs a thesaurus using the measure on a parsed corpus. Lexicographic glosses (Pedersen et al. 2004) and surrounding contexts (Pantel et al. 2009) can also be used for feature representation in word similarity estimation. More recently, graph based methods have also be developed for measuring semantic similarity and relatedness between terms. Jarmasz and Szpakowicz (2003) calculate the semantic distance of two words according to the length of path between the two words in the Roget's thesaurus. Agirre et al. (2009) apply personalized PageRank over WordNet graph for a pair of words, producing a probability distribution over WordNet synsets, and then compare these two distributions by encoding them as vectors and computing the cosine value between the vectors. Strube and Ponzetto (2006) integrate path based measures, information content based measure and text overlap based measures over Wikipedia for computing semantic relatedness. The major drawback of semantic relatedness estimation is that it does not define a particular kind of relation. For example, "*desk*" is related to "*chair*", and "*Apple*" is related to "*Steve Jobs*". However, the two relations are not the same type.

*2.3. Word Sense Disambiguation*

The concept disambiguation problem in the coordinate term extraction task is similar to the word-sense disambiguation (WSD) task, which is a process of deciding the particular meaning of words in context. WSD has been described as an AI-complete problem (Mallery 1988), i.e. its difficulty is equivalent to solving central problems of artificial intelligence, e.g. the Turing Test (Turing 1950).

The particular WSD problems are formalized differently due to fundamental questions, the granularity of sense inventories, the set of target words to disambiguate, etc. Supervised techniques view WSD as a task of classifying word mentions into the "fixed-list of senses" (Galley and McKeown 2003; Màrquez et al. 2006; Navigli and Velardi 2005), and unsupervised WSD tries to induce word senses directly from the corpus by grouping together similar examples (Pantel and Lin 2002; Purandare and Pedersen 2004). The major disadvantage of supervised techniques is that it is usually very hard to get the complete word sense definitions and a large annotated corpus for all the words. The major drawback of unsupervised methods is that the set of obtained senses is unpredictable, and may not be compatible to existing taxonomies. The WSD task can also be distinguished as *lexical sample* (*targeted WSD*) and *all-words WSD*. The *lexical sample WSD* systems are required to disambiguate a restricted set of target words, usually one word per sentence. The *all-words WSD* systems are expected to disambiguate all open-class words in a text (Mihalcea 2005; Zhong and Ng 2009).

The central problems in WSD are the feature representation and similarity calculation. Several kinds of features have been used in this task, including local context features, topical features, syntactic features and semantic features. Schütze (1998) uses the vector of close neighbors in the corpus to represent a word, and measures the similarity by the cosine value between two vectors. Lin (1998) extracts the dependency triples from the text corpus, describe a word with the frequency counts of all the dependency triples in which the word is the first element, and calculate the similarity between two words using the mutual information. Veronis (2004) builds co-occurrence graphs, and uses minimum spanning tree to disambiguate specific instance of a target word. Mihalcea et al. (2004) propose to use semantic graph for WSD. The method builds a graph that represents all the possible senses of words in a text and interconnects pairs of senses with meaningful relations. After applying PageRank on the graph, the highly ranked sense of each word in context is chosen.

## 3. Problem definition

In this section, we first give an explanation of the involved terminology, and then discuss the task of coordinate term extraction.

A **term** is a character sequence which has a specific meaning in nature language. It refers to something in the world, either concrete or abstract. For example, 北京/*Beijing* is a term which refers to an area on the earth. *Science* is another term which refers to a particular kind of thoughts and activities. Note that the mapping function between terms and things is not bijective. A term may refer to different things. For example, 苹果/*apple* can refer to a kind of fruit, or a computer company (*Apple Inc.*). 华盛顿/*Washington* can refer to a city (华盛顿特区/*Washington DC*), a state of US (华盛顿州/*state of Washington*), or a person (乔治华盛顿/*George Washington*, etc.). A thing may be represented by several terms. For example, 北京/*Beijing* and 北平/*Peping* both refer to the same city. 飞机/*airplane* has the same meaning with 固定翼飞行器/*fixed wing aircraft*.

The things in the world are related, and can be organized as a taxonomy. In the taxonomy, things in common are grouped into sets. Each set is called a **concept**, and every individual thing within this set is called an **instance** of the concept. Both the concepts and the instances can also represented by terms. For example, 城市/*city* refers to a concept which represents a relatively large and permanent settlement, and 北京/*Beijing* is an instance of 城市/*city*. A subset of instances of a concept may compose another concept. For example, the set of cities which locate in China can be called 在中国的城市/*city in China* or 中国的城市/*Chinese city*. In this case, the subset is called a **sub-concept** of the original concept. Note that an individual thing can be an instance of different concepts simultaneously. For example, 金门大桥/*Golden Gate Bridge* is an instance of 旧金山旅游景点/*tourist destination in San Francisco*, while it is also an instance of 世界著名桥梁/*famous bridges in the world*.

Formally, let $con_i = \{ind_j\}$ denote a concept, where each $ind_j \in con_i$ is an individual things and also an instance of $con_i$. The concept $con_j$ is a sub-concept of $con_i$ if $con_j \subset con_i$.

The concept/instance taxonomy can be represented by a hierarchical graph, as illustrated in fig. 2. In the graph, each concept or individual thing is regarded as a node. To simplify the graph, a concept is only linked with its immediate sub-concepts and instances. The concept that contains the sub-concepts and instances is placed in the upper level, while the sub-concepts and instances are placed in the lower level. In each pair of linked nodes, the semantic extension of the lower one (called **hyponym**) is contained by the semantic extension of the upper one (called **hypernym**).

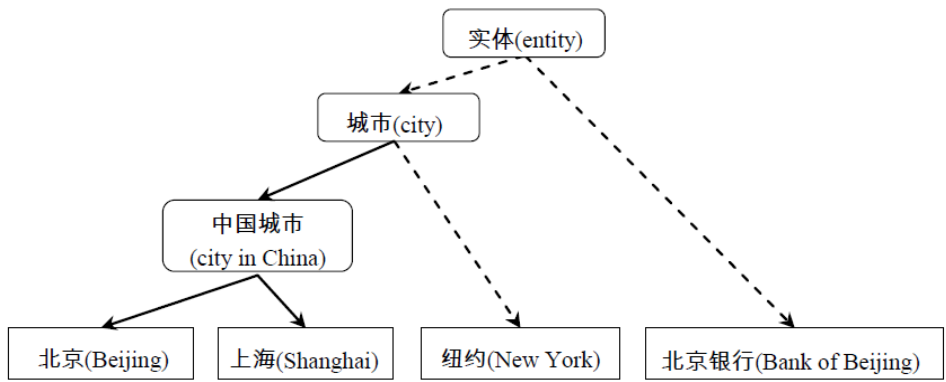

**Fig. 2**: A sample taxonomy hierarchy of concepts and instances (the rounded rectangles denote concepts, the rectangles denote instances, the links denote the "hypernym-hyponym" relations, and the dotted links mean that some concepts are omitted in the path)

Formally, the taxonomy graph is defined as $G = \{V, E\}$. $V = \{con_i\} \cup \{ind_j\}$ is the set of concepts and individual things. $E = E_{cc} \cup E_{ci}$ defines the relations between nodes. $E_{cc}$ represents the hypernym/hyponym relations between concepts

$$E_{cc} = \left\{ <con_i, con_j> | con_j \subset con_i \land \forall con_k \left( con_j \subset con_k \rightarrow con_i \subseteq con_k \right) \right\}$$

$E_{ci}$ represents the instance-of relations between individual things and concepts.

$$E_{ci} = \{<con_i, ind_j> | ind_j \in con_i \land \forall con_k (con_k \subset con_i \to ind_j \notin con_k)\}$$

If there is a path from $con_i$ to $con_j$ (or $ind_j$) in the graph, then $con_i$ is a hypernym of $con_j$ (or $ind_j$), and $con_j$ (or $ind_j$) is a hyponym of $con_i$. The length of the path (i.e. the number of edges in the path) can reflect the degree of generalization of the hyponym with respect to the hyponym. If the path is long, then $con_i$ is a general concept for $con_j$ (or $ind_j$).

The **coordinate relation** refers to the relations between things that have same hypernym in the taxonomy, i.e.

1. Instances of the same concept, or
2. Concepts which are hyponyms of the same concept.

Note that all things can be instances of the root concept (i.e. *object*), however, usually only instances of a specific concept are considered as coordinate. For example, 北京银行/*Bank of Beijing* is usually not considered as a coordinator of 北京/*Beijing*, because their common hypernym is too general, and the semantic relatedness between them is too weak.

If two things are coordinate, they are also called **coordinators** of each other. For example, 北京/*Beijing* and 上海/*Shanghai* are coordinate, because they are both instances of 中国城市/*city in China*. 城市/*city* is not a coordinator but a hypernym of 北京/*Beijing*. In reality, a term can usually be replaced with its coordinators while keeping the result grammatically correct (but may not be truth). For example,

> 北京面积比香港大。/ ***Beijing** is larger than Hong Kong.*
> 上海面积比香港大。/ ***Shanghai** is larger than Hong Kong.*
> 城市面积比香港大。/ ***City** is larger than Hong Kong.*

The second sentence is obtained by replacing 北京/*Beijing* with 上海/*Shanghai*. It is grammatically correct (and is also truth). The third sentence is obtained by replacing 北京/*Beijing* with 城市/*city*. This sentence is not grammatically correct any more, because 城市/*city* and 香港/*Hong Kong* do not belong to the same concept and thus they cannot be compared with each other.

Formally, two individual things $ind_i$, $ind_j$ are coordinate with each other iff

$$\exists con_k (P(con_k, ind_i) \neq \varnothing \land P(con_k, ind_j) \neq \varnothing \land |P(con_k, ind_i)| < \varepsilon \land |P(con_k, ind_j)| < \varepsilon)$$

where $P(con_k, ind_i)$ and $P(con_k, ind_j)$ denote the paths from $con_k$ to $ind_i$ and $ind_j$, respectively; $|P(con_k, ind_i)|$ and $|P(con_k, ind_j)|$ denote the length of the paths; $\varepsilon$ constrains the relatedness between $ind_i$ and $ind_j$ by constraining the degree of generalization of their common hypernym $c_k$. Similarly, two concepts $con_i$, $con_j$ are coordinate with each other iff

$$\exists con_k (P(con_k, con_i) \neq \varnothing \land P(con_k, con_j) \neq \varnothing \land |P(con_k, con_i)| < \varepsilon \land |P(con_k, con_j)| < \varepsilon)$$

Because concepts/instances can be represented by terms, we can also define coordinate relations between terms. Terms are **coordinate** with each other iff their represented individuals/concepts are coordinate. In the rest of this paper, we will no longer distinguish a concept/instance with its representing term unless emphasized particularly.

What calls for special attention is that the coordinate relation between terms is not transitive, i.e. two terms may not be coordinate with each other even they are both coordinate with the third term. This is caused by two kinds of ambiguity.

First, a term may refer to several concepts/individual things, and each concept/individual thing will lead to different coordinators. For example, 苹果/*apple* can mean a kind of fruit, or an IT company. The coordinate terms of 苹果(水果)/*apple (fruit)* include 橘子/*orange*, 芒果/*mango*, etc. The coordinators of 苹果(公司)/*apple (company)* consist of 微软/*Microsoft*, 谷歌/*Google*, etc. Obviously, 橘子/*orange* and 微软/*Microsoft* are not coordinate, since the underlying concepts are not related at all.

Second, an individual thing can be the instance of several concepts simultaneously, and thus it can result in different coordinators. For example, when considering 奥巴马/*Obama* as one of the US Presidents, 华盛顿/*Washington*, 亚当斯/*Adams* and 杰弗逊/*Jefferson* are the coordinate terms. When considering 奥巴马/*Obama* as one of the current heads of government, 普金/*Putin*, 卡梅伦/*Cameron* and 奥朗德/*Hollande* are its coordinate terms. It is usually not appropriate to mix up these two lists, otherwise the underlying concept will be overgeneralized.

In order to overcome the naturally ambiguity of a term, several strategies can be used. The simplest one is asking for an explicit sense for the term, e.g. requiring the user to provide its hypernym. The second strategy is requiring more terms as input, and taking the overlapped concept as the exact sense. For example, if the user inputs 奥巴马/*Obama* and 克林顿/*Clinton*, then we can imply that he/she wants a list of the US Presidents. These two strategies both need extra information. As the third strategy, the system can find all possible coordinate terms, organize them according to the underlying concepts, and let the user decide which is he/she actually needs. For example, if the user issues 华盛顿/*Washington* as the seed, the system may respond with the following results:

- 亚当斯/*Adams*, 杰弗逊/*Jefferson*, 麦迪逊/*Madison*, 门罗/*Monroe*, ...    (*US President*)
- 纽约/*New York*, 芝加哥/*Chicago*, 洛杉矶/*Logs Angles*, 旧金山/*SanFrancisco*, ...    (*City*)
- 加利福尼亚/*California*, 佛罗里达/*Florida*, 俄勒冈/*Oregon*, 德克萨斯/*Texas*, ...    (*State*)
- 伦敦/*London*, 巴黎/*Paris*, 柏林/*Berlin*, 罗马/*Rome*, ...    (*Capital*)

Based on previous definitions and discussions, we can now define our task. The coordinate term mining task aims to extract terms which are coordinate with each other. According to different input, it can be formalized as different sub-tasks. In this study, we focus on the subtask of extracting the coordinators of a user given term (called **seed term** or **query term**) in Chinese from the web as follows:

*The user issues a single seed term $t_s$ (e.g. 苹果/Apple), and the task aims to find one or more lists of other probable terms (e.g. {香蕉/Banana, 橘子/Orange, 芒果/Mango...}, {微软/Microsoft, 谷歌/Google, 雅虎/Yahoo, ...}), where each term in the results is coordinate with the seed term, and coordinate with other terms in the same list. In addition, the terms in each list is ranked according to their relatedness to the seed term.*

The reason of term ranking is that the users usually pay more attention to the top few results, and thus it is more important to guarantee those results correct.

## 4. Proposed algorithm

*4.1 Overview*

It is easy to figure out whether terms are coordinate if the corresponding taxonomy hierarchy is known. Many efforts have been made to build such knowledge bases, however they are still far from completion. Therefore, additional knowledge and resources are required for coordinate term extraction.

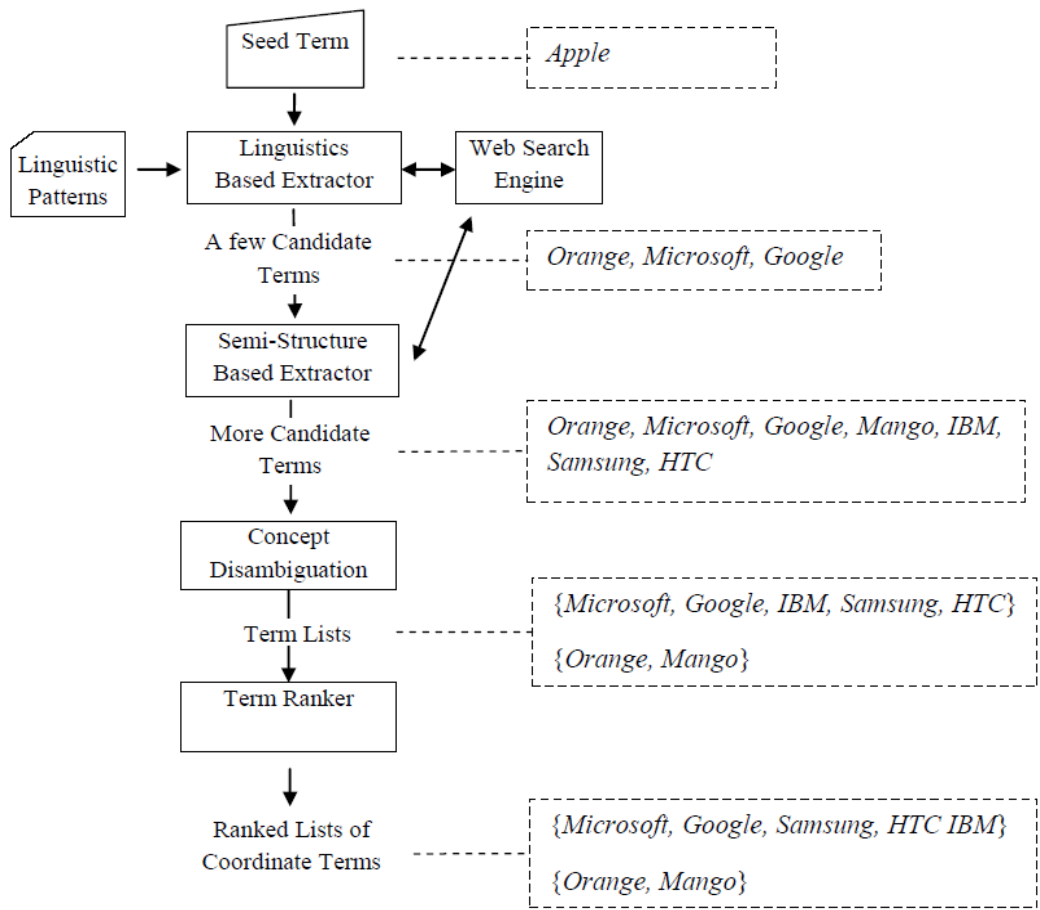

**Fig. 3**: The framework of coordinate term mining system

Being the largest public accessible corpus, the Web contains almost all kinds of knowledge. However, the web pages lack of semantic labels, and thus particular techniques are required to mine useful information from the mass data.

Intuitively, there are two empirical findings of appearances of coordinate terms. First, coordinate terms usually co-occur in comparative and coordinative sentences. These sentences usually follow particular linguistic patterns, and thus it is not hard to identify these sentences and extract the corresponding coordinate terms. Second, coordinate terms are usually organized in some structures (such as lists and tables) in web pages. Many of these structures in web pages are formatted by templates, and thus they can be extracted by automatically learned patterns according to their repeating appearances.

In this study, we propose a system named CTMS (Coordinate Term Mining System) to make use of these two findings for extracting coordinate terms. The input of CTMS is one single term, which will alleviate the user's search burden. The novelty of CTMS is summarized as follows:

- CTMS analyzes natural language texts and semi-structured HTML pages to get candidate coordinate terms for a single given term.
- CTMS employs the clustering technique to group terms of different concepts and improve the intelligibility of mined results.

The framework of our system is shown in fig. 3. The example result for each step is attached in the dashed rectangle. Our system consists of five steps: the step of initial candidate set construction extracts an initial set of coordinate terms from Web search results using only the single query term and linguistic patterns; the step of candidate set expansion extracts more candidate terms from web pages using set expansion techniques; the concept disambiguation step groups the results of different concepts into different lists; the term ranking step ranks obtained terms according to their relatedness with the query term using a random graph walk algorithm.

Note that the proposed system relies on a Web search engine to retrieve search results in the first two steps. In this study, we adopt the widely used Google Search[2], but other search engines will work as well.

In the following subsections, we will describe the above steps in more details, respectively.

*4.2 Initial Candidate Set Construction*

Given a single seed term $t_s$, this step aims to find a few candidates of coordinate terms through natural language text analysis. Because the candidate set will be used as the input for the next set expansion step, the precision is the first priority. This step is based on the top 200 snippets (including titles) returned by Google Search.

In Chinese, there are two kinds of sentence structures that may contain coordinate terms. The first kind is comparison, which is used to describe the relative positions of two or more objects. The compared objects in the comparative sentences need to have something in common, and thus they are likely to be coordinate terms. For example,

a) 为什么大街上**宝马**比**奔驰**多？(*Why are there more **BMW** than **Benz** in the street*?)

This sentence makes a comparison between two coordinate terms 宝马 (*BMW*) and 奔驰 (*Benz*).

The second kind is coordination, which links together two or more elements of the same grammatical form. If the constituents are terms, then they are likely to be coordinate. For example,

b) 三星和**LG**难以短期内超越诺基亚。(*It is hard for **Samsung** and **LG** to surpass Nokia in a short time*.)

In this sentence, there is a coordination between two coordinate terms 三星 (*Samsung*) and *LG*. Actually this sentence is also a comparative sentence, which is indicated by the verb 超越 (*surpass*). Thus the term 诺基亚 (*Nokia*) is also coordinate with the terms 三星 (*Samsung*) and *LG*.

In order to find such structures, we may search the web using the given seed term, and check each sentence in the retrieved results. However, such structures do not occur frequently, and thus it is possible that none can be found in the top results. In Chinese language, there are several patterns of comparison (Che 2005) and coordination (Qiang et al. 1999). Most patterns contain some function words, such as 比 (*than*) and 和 (*and*) in the above two examples. These words are grammaticalized, and they occur much more frequently in comparisons and coordinations than in other kinds of sentences. Based on this phenomenon, we can use these function words together with the user given seed term as queries to the search engine, and hopefully the search engine will retrieve more comparisons and coordinations.

In order to extract terms from comparisons and coordinations, we need to detect the exact boundary of those terms. The function words define one of the boundaries naturally, but the other boundary is not easy to confirm. Unlike the western languages, there is no mark (such as space and capitalization) between terms in Chinese sentences. In addition, because of the complexity and irregularity of web texts, traditional Chinese word segmentation, entity recognition and syntactic parsing techniques usually do not perform well on the web corpus. Thus we cannot rely on those methods either.

Note that there is more than one way to express the comparison and coordination between two terms. We can always rewrite a comparison by swapping the positions of compared terms and altering the comparison result accordingly. For example, the example a) can be rewritten as:

---

[2] http://www.google.com

c) 为什么大街上**奔驰**比**宝马**少? (*Why are there less **Benz** than **BMW** in the street*?)

In example a), the function word 比 (*than*) defines the left boundary of the term 奔驰 (*Benz*), while in example c) it defines the right boundary of this term. Thus we can extract the term by intersecting the tokens (i.e. character) before the function word and the tokens after the function word. The coordinations are the same. We can usually swap the positions of two constituents without changing the meaning. For example, the example b) can be rewritten as

d) **LG**和**三星**难以短期内超越诺基亚。(*It is hard for **LG** and **Samsung** to surpass Nokia in a short time*.)

In order to find the terms occurring both before and after the function words, we issue two queries for each indicating function word *f*, i.e. "*f* + $t_s$" and "$t_s$ + *f*". For example, if the seed term is 宝马 (*BMW*), and the function word is 比 (*than*), then the two queries are "宝马比" ("*BMW than*") and "比宝马" ("*than BMW*"), respectively.

The extraction procedure is described in fig. 4. We first submit queries to the search engine and gather retrieved snippets and titles (line 2-5). Note that the quotation marks in the queries are required to search for exact phrases. The token sequences that occur before and after the function words are then extracted and scored according to their frequencies (line 6-8). Those whose scores are larger than a threshold $\tau$ ($\tau = 2$ in this study) are selected and ranked (line 9-10). Finally the best *N* (*N* = 5 in this study) results are extracted as candidate terms that are likely coordinate with the seed term (line 12-14).

```
Input:    seed term t_s, indicating function word set F
Output:   C_init - sets of potential coordinators of t_s
Procedure:
1  Sentence set S ← ∅
2  for each indicating functor f in F
3      submit query "t_s·f" to web search engine, add sentences in results to S
4      submit query "f·t_s" to web search engine, add sentences in results to S
5  end for
6  for all possible token sequence x
7      n_x = |{s∈S | s = s_p·x·f·t_s·s_s}|     -- dot(·) means string connection
8      m_x = |{s∈S | s = s_p·t_s·f·x·s_s}|
9      score(x) = n_x * m_x
10     if score(x) > τ, then L ← L ∪ {x}
11 end for
12 sort L in descending order by score of each token sequence
13 C_init ← Top N of L
14 return C_init
```

**Fig. 4**: The linguistics based initial coordinate term extraction algorithm

*4.3 Candidate Set Expansion*

This step aims to expand the initial candidate set $C_{init}$ into a more complete set $C_{expand}$. We note that the initial candidate set is constructed based on text analysis of titles and snippets in the search results. Though the precision of $C_{init}$ is high, the recall is usually low because those linguistic patterns do not appear frequently in the web pages. In the meantime, there are many other clues in the semi-structured web pages for extracting coordinate terms. For example, a list of items or entities are usually embedded with HTML tags such as "*<li>*" and "*</li>*", or "*<td>* and *</td>*". We believe that items or entities within the same classes will appear in similar formatting structures on the same web pages. Thus, the characteristics of semi-structured web pages can be exploited to find more coordinate terms for a few initial candidate terms.

In order to retrieve the pages which contain such lists of terms, we must submit elaborate queries to the web search engine. When we query with the single seed term, the web search engine tends to return descriptive articles such as instructions and reviews, which usually do not contain any lists. For example, fig. 5a shows the top five results of query

*BMW*. All of them talk about the single object *BMW*, and actually none of them contain any list of coordinate terms of *BMW*.

Intuitively, if a page contains a few coordinate terms, it is possible to contain more coordinate terms. For example, fig. 5b shows the top five results of query "*BMW Ferrari*". We can easily figure out that several of them contain lists of terms coordinate with *BMW* (e.g. *Porsche*, *Jaguar*, *Audi*, etc.). Based on this assumption, we can use the given seed term and a few coordinate terms together as queries, and hopefully the search engine will prefer pages that contain lists of terms, e.g. navigation pages. Note that some coordinate terms are already available in section 4.2. In this paper, we call the collection of the seed term and initial candidate set as the **extended seed term set** $T_{SE} = \{t_s\} \cup C_{init}$, and a term in the extended seed set is called an **extended seed term**.

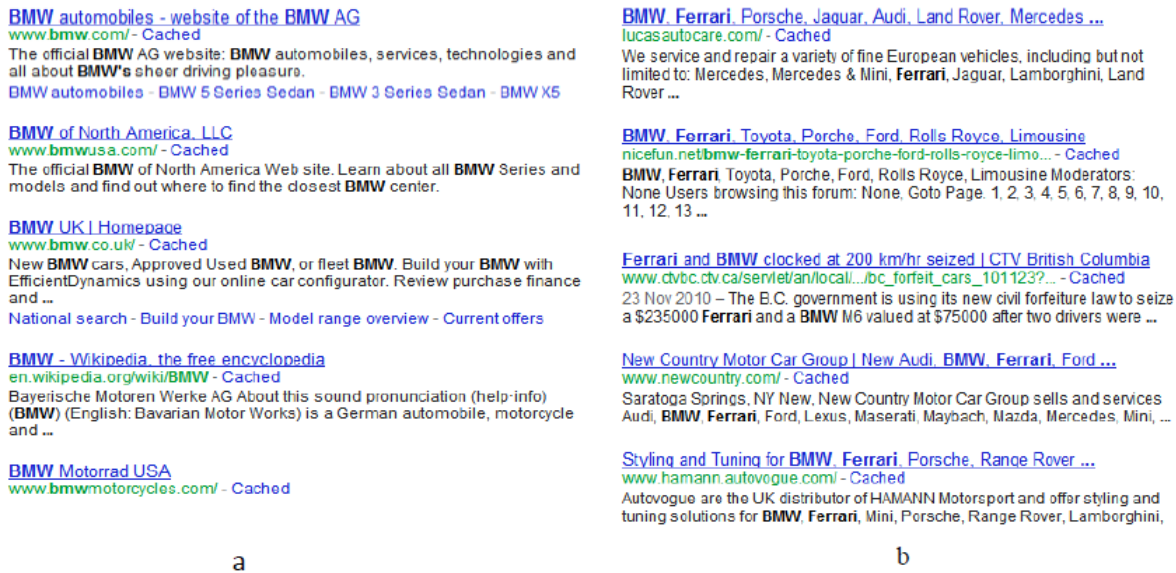

**Fig. 5**: a) Top five search result of the query *BMW* b) Top five search result of the query *BMW Ferrari*

In modern web page design, web lists are usually generated by applying regular templates on each item successively. It means that the items in the list share similar HTML tags, and lie at the similar positions in the HTML structure tree. The HTML structure can be represented using a DOM (Document Object Model) tree. Each node in the DOM tree represents an HTML element. A parent element is linked with all its immediate child elements, i.e. the elements which are contained in the parent element, but not contained by another element. The DOM path of a node is the sequence of nodes from the root node down to the node itself. For example, fig. 6a shows a list of computer brands in an IT web site. The corresponding HTML source code snippet is shown in fig.6b, and part of the DOM tree is shown in fig.6c. As we can see, the items share the same contextual code (*.shtml">…</a></div>*), and the same DOM path (*…div/div/div/a/span/#text*).

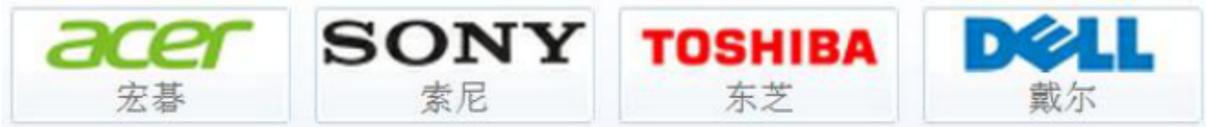

```
<div class="cur_dh brand">
    <div class="dh border menu_div" id="menu_1">
        <div class="brand_logo clearfix">
            <a href="http://ideapad.zol.com.cn/" class="all_logo" target='_blank'>
                <img src="http://digital.zol-img.com.cn/226_module_images/12/4e7321d8bf978.gif" width="91"
                height="20" />
                <span>宏碁</span>
            </a>
            <a href="http://haseebbs.zol.com.cn/" class="all_logo" target='_blank'>
                <img src="http://digital.zol-img.com.cn/226_module_images/13/4e73205557f04.jpg" width="91"
                height="20" />
                <span>索尼</span>
            </a>
            <a href="http://hpbbs.zol.com.cn/" class="all_logo" target='_blank'>
                <img src="http://digital.zol-img.com.cn/226_module_images/13/4e73209b295a3.jpg " width="91"
                height="20" />
                <span>东芝</span>
            </a>
            <a href="http://asusbbs.zol.com.cn/" class="all_logo" target='_blank'>
                <img src="http://digital.zol-img.com.cn/226_module_images/13/4e7320fa8700f.jpg" width="91"
                height="20" />
                <span>戴尔</span>
            </a>
        </div>
    </div>
</div>
```

b)

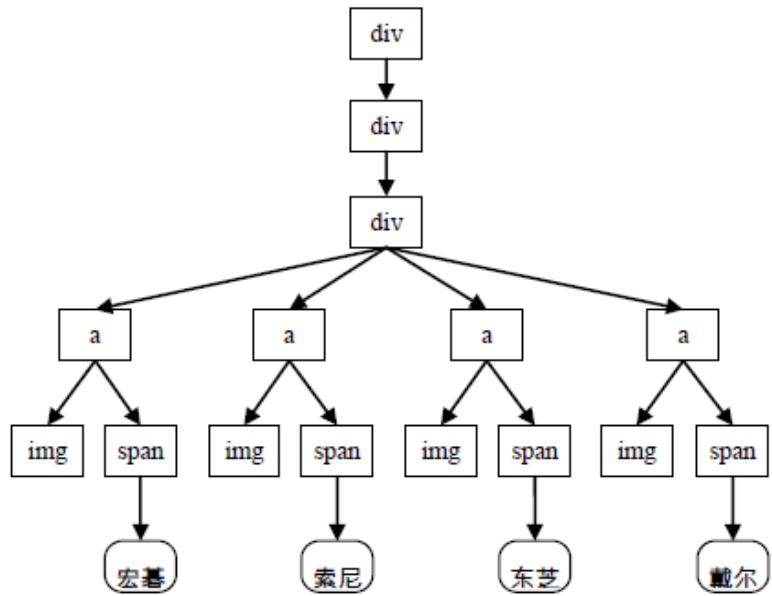

c)

**Fig. 6**: a) A list of computer brands b) Fragment of the source code c) Fragment of the DOM tree

Definition:
1. $t_{i,j} = j^{th}$ occurrence of extended seed term $t_i \in T_{SE}$ in the web page $d$
2. $l_{i,j}$ and $r_{i,j}$ is the left and right context of $t_{i,j}$ respectively, i.e. $d = l_{i,j} \cdot t_{i,j} \cdot r_{i,j}$.
3. $TL_l = \{t_{i,j} | l \text{ is left context of } t_{i,j}\}$, $TR_r = \{t_{i,j} | r \text{ is right context of } t_{i,j}\}$
4. common left contexts $CL(T) = \{l | \exists t_{i,j}, t_{p,q} \in T: (i \ne p \wedge t_{i,j}, t_{p,q} \in TL_l)\}$
5. longest common left contexts $LCL(T_{SE}) = \{l | l \in CL(T) \wedge \nexists l' \in CL(T): l \text{ is suffix of } l'\}$
6. common right contexts $CR(T) = \{r | \exists t_{i,j}, t_{p,q} \in T: (i \ne p \wedge t_{i,j}, t_{p,q} \in TR_r)\}$
7. longest common right contexts $LCR(T) = \{r | r \in CR(T) \wedge \nexists l' \in CR(T): r \text{ is suffix of } r'\}$
8. $p_{i,j} = <tag_{i,j,1}, ... tag_{i,j,m}>$ is the DOM path of $t_{i,j}$, where $tag_{i,j,k}$ is the tag name
9. $P = \{p_{1,1}, ..., p_{1,n1}, ..., p_{k,1}, ..., p_{k,nk}\}$
10. $TP_p = \{t_{i,j} | p_{i,j} = p\}$

Input: A set of extended seed terms $T_{SE}$, a Web page $d$.
Output: Set of wrappers $W$

LearnWrapper algorithm:
```
1    let wrapper set W ← ∅
2    for each p ∈ P
3        for each l in LCL(TP_p)
4            for each r in RCL(TL_l ∩ TP_p)
5                w ← <l, r, p>
6                if w satisfy the constraints
7                    W ← W ∪ {w}
8                end if
9            end for        -- each r
10       end for            -- each l
11       for each r' in RCL(TP_p)
12           for each l' in LCL(TR_r' ∩ TP_p)
13               w ← <l', r', p>
14               if w satisfy the constraints
15                   W ← W ∪ {w}
16               end if
17           end for        -- each l'
18       end for            -- each r'
19   end for                -- each p
20   return W
```

**Fig. 7**: The semi-structural wrapper learning algorithm

Given a web page and an extended seed set, we can learn wrappers by finding the common context of the extended seeds, and then use these wrappers to extract more terms. Formally, a **wrapper** $w$ is a triplet $<l, r, p>$, where

- $l$ is the left contextual HTML code of the term (i.e. the tokens immediately before the term),
- $r$ is the right contextual HTML code of the term (i.e. the tokens immediately after the term),
- $p$ is the path of the DOM node where the term occurs.

A wrapper $w$ is adopted if it bracket at least two occurrences of different extended seed terms and there is no superior wrapper $w'$ that matches the same term occurrences as $w$. We adapt the learning algorithm in (Wang and Cohen 2007) to find such wrappers in a single page, as illustrated in fig. 7. We first group the instances of extended seed terms according to their DOM paths, and then find the common left and right contexts in each group. Note that a common left/right context needs to match occurrences of two or more extended seed terms in our algorithm, but it is not required to match occurrences of all extended seeds as in (Wang and Cohen 2007). This looseness can lead to more wrappers without hurting precision too much. In line 6 and 11, we use several heuristic rules to wipe out noisy wrappers. Basically, a good wrapper should meet:

1. Either *l* or *r* is not white space;
2. Both *l* and *r* are punctuation, or neither *l* nor *r* is punctuation;
3. If *l* and *r* are not punctuation, then $|l|+|r| \geq \kappa$ ($|l|$ and $|r|$ are the lengths of *l* and *r*, respectively. $\kappa = 4$ in this study).
4. *p* ends with a textually node, i.e. an attribute node or a text node.

Note that the wrappers are highly dependent on the seed terms and the web pages, which means that they can only be applied on the pages where they are learned. A character sequence *c* is extracted if it occurs in a DOM node whose path is matched, and it is bracketed by the context in the wrapper, (i.e. $\exists x \exists y(d=x \cdot l \cdot c \cdot r \cdot y \wedge p_c = p)$, where *x* and *y* are source code strings, and $p_c$ is the DOM path of *c*). This task can be achieved by finding the occurrences of left and right contexts using multiple patterns matching algorithms (Aho and Corasick 1975), and then checking the DOM paths of the bracketed texts. The extraction algorithm is illustrated in fig. 8.

```
Input: web page d, wrapper set W
Output: set of potential coordinators C_expand(d)
Definition:
1.  Wl = {l | ∃r ∃p: <l, r, p> ∈ W}
2.  Wr = {r | ∃l ∃p: <l, r, p> ∈ W}
3.  R(l) = {r | ∃p: <l, r, p> ∈ W}
4.  L(r) = {l | ∃p: <l, r, p> ∈ W}
5.  P(l, r) = {p | <l, r, p> ∈ W}
6.  LeftPos(r) = {<l, pos> | l ∈ L(r) ∧ substring(d, pos, |l|) = l}
7.  Path(pos) is the DOM path of node that contains the pos^th character in d
8.  substring(str, pos, length) get a substring of length characters from str starting at pos
9.  AhoCorasick : <{str}, d> → {<str, pos>} is string matching function, where str is a string, pos is
    the possition of a occruence of str in document d. The results are sorted by ascending order of
    pos

ExtractTerm algorithm:
1   C_expand(d) ← ∅
2   for each match <m, pos> in AhoCorasick(Wl ∪ Wr, d)
3     if m ∈ Wl
4       for each r ∈ R(m)
5         LeftPos(r) ← LeftPos(r) ∪ {<m, pos>}
6       end for
7     end if
8     if m ∈ Wr
9       for each <l, pos'> ∈ LeftPos(m)
10        if Path(pos-1) = Path(pos'+|l|) ∧ Path(pos-1) ∈ P(l, m)
11          C_expand(d) ←C_expand(d) ∪ substring(d, pos'+|l|, pos – pos' - |l|)
12        end if
13      end for
14    end if
15  end for
16  return C_expand(d)
```

**Fig. 8**: The wrapper based single page coordinate term extraction algorithm

Overall, the procedure of the semi-structure based expansion algorithm is illustrated in fig. 9. In line 3-4, we do not use the whole extended seed term set as the query to the web search engine, otherwise the search engine will return very few results. Instead, we select a few extended seed terms at each time, and repeat until all possible selections have been used. However, in line 6, we still use the whole extended seed set $T_{SE}$ rather than the subset *S* to learn wrappers.

```
Input:  extended seed term set T_SE
Output: set of potential coordinators of seed term C_expand
Procedure:
1    C_expand ← ∅
2    do
3        select T ⊂ T_SE
4        submit T to web search engine
5        for each retrevied web page d
6            W ← LearnWrapper (T_SE, d)
7            for each wrapper w ∈ W
8                C_expand ← C_expand ∪ ExtractTerm (d, w)
9            end for    -- each w
10       end for    -- each d
11   loop till all selections have been tried
12   return C_expand
```

**Fig. 9**: The semi-structure based coordinate term expansion algorithm

*4.4 Concept disambiguation*

In our task, the user inputs only one seed term. This term may have several meanings. It is impossible to infer which meaning the user actually refers to because no contextual information is given. Hence the terms extracted in the previous steps can be coordinate with different meanings of the seed term and thus belong to several concepts respectively. For example, the results of 华盛顿 (*Washington*) can include 杰弗逊 (*Jefferson*) and 纽约 (*New York*). In this case, it will cause confusions if these terms are mixed together. This step aims to group the terms into different sets that represent different concepts.

Since most lists in the web pages are created and edited manually, each web list should describe a certain semantic meaning (either a general concept, e.g. *US Presidents*, or a underlying concept, e.g. *Information of Washington*), and thus it is reasonable to assume that terms within each list belong to the same concept. If two web lists contain similar terms, they are likely to be semantically similar with each other. Note that if a web list is totally contained by another web list, then the concept which the contained list represents is very likely be a hyponym of the concept which the container list represent. In this case we also group them together. Besides, if two web lists are semantically related, it is highly possible that the contextual texts of the terms belong to the similar topic. Conversely, the similarity of two web lists' contextual texts can also indicate to the semantic similarity of those web lists. Based on these two assumptions, we can calculate the semantic similarity of two web list by taking into account their contents and contexts, and then group the similar web list to obtain the latent concepts.

Formally, let $C = \{c_1, c_2, \ldots c_n\}$ be the collection of all candidate terms. A web list $wl_i \subset C$ is a small collection of candidate terms extracted using a wrapper from page $d_j$. We represent $wl_i$ as a binary vector of term $wl_i = <o_{i1}, o_{i2}, \ldots o_{in}>$, where $o_{ij} \in \{0, 1\}$ represents whether the term $c_j \in C$ is contained in $wl_i$. We use the text within a window around $wl_i$ in $d_j$ to represent the context of $wl_i$, and model it as a vector of words $ct_i = <w_{i1}, w_{i2}, \ldots w_{im}>$, where $w_{ik}$ is the *tf·idf* weight of the word $word_k$ :

$$w_{ik} = tf(word_k, d_j) \cdot \log\left(\frac{|B|}{\sum_{d' \in B} I(word_k, d') + 1}\right)$$

where $tf(word_k, d_j)$ is the frequency of the occurrences of $word_k$ in the document $d_j$; $B$ is a large background corpus; $|B|$ is the number of documents in $B$; and $I(word_k, d')$ is an indicator function of whether $word_k$ occurs in $d'$. The similarity of two web lists is calculated as follows:

$$Sim(wl_i, wl_j) = \frac{\lambda \sum_{k=1}^{n} o_{ik} o_{jk}}{\min\left(\sum_{k=1}^{n} o_{ik}, \sum_{k=1}^{n} o_{jk}\right)} + \frac{(1-\lambda)\sum_{k=1}^{m} w_{ik} w_{jk}}{\sqrt{\sum_{k=1}^{m} w_{ik}^2} \cdot \sqrt{\sum_{k=1}^{m} w_{jk}^2}}$$

The first part of the above equation is the similarity of lists' contents, and it will be higher if most terms in one list are included by another list. The second part is the similarity of contexts. $\lambda \in [0, 1]$ is a parameter that tunes the importance of content-based similarity and the context-based similarity. In this study, we set $\lambda = 0.5$.

In this study, we use the agglomerative hierarchical clustering algorithm (Jain and Dubes 1988) to group the web lists into different concepts. The web lists are merged till their similarities are lower than a threshold (set as 0.65 in this study). The clusters which do not contain the seed term are discarded because the relationships between the seed term and these clusters are weak. In addition, because of the complicated web page contents and layouts, it is inevitable that some errors are also extracted by the mining algorithm. The erroneous terms are usually page-dependent so they do not appear in many other pages. Hence these erroneous lists usually do not merge with other lists and they form dispersive small clusters. Therefore, we introduce a minimal support for the clusters to wipe out the erroneous clusters, i.e., a cluster containing fewer than $\eta \cdot |L|$ web lists is discarded, where $L$ is the collection of all web lists, and $\eta=5\%$ in this study.

*4.5 Coordinate term ranking*

This step aims to sort the extracted candidate terms according to their saliency scores. Among the candidate terms, some of them are strongly related to the seed term, while some of them are noises. When returning the user with a list of coordinate terms, it is preferable to place the most important terms at the top, and leave the least important terms at the bottom.

Intuitively, there are two evidences of terms' saliency scores. The first evidence is distributions of the terms. Similar to the HITS model for web page ranking, if a term appears in many good web lists, then it is very likely to be a salient term. On the other hand, if a web list contains many significant coordinate terms, then it is likely to be good. The second evidence is the linguistic clues. Coordinate terms sometimes have similar word-formation such as preceding word or following word. For example, we can infer that "*Peking University*" and "*Stanford University*" are significant coordinators because they share the word "*University*".

To utilize these clues, we build a graph to represent the relations among term candidates, web lists and preceding/following words within each concept. As illustrated in fig. 10, each candidate term is connected with the web list which contains it, and the preceding/following words which it contains. Formally, let $C_c = \{c_i\}$ be the extracted candidate term set in a concept, $L_c = \{wl_j\}$ be the web list set in the concept, and $F_c=\{pf_k\}$ be the preceding/following word set of terms in $C_c$, then the graph can be denoted as $G=<V, E>$, where $V = C_c \cup L_c \cup F_c$, and $E = \{(c_i, wl_j) \mid wl_j$ contains $c_i \} \cup \{(c_i, pf_k) \mid c_i$ contains $pf_k\}$.

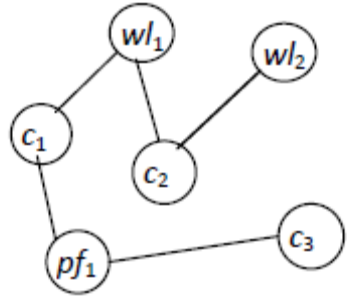

**Fig. 10**: A example of the relation graph (*c, wl, pf* denotes candidate term, web list and preceding/following word, respectively)

We apply the random walk with restart (RWR) algorithm (Pan et al. 2004) on each graph to calculate the saliency values of candidate terms. In the graph, we start with the vertex of user given seed term $v_0$. At each step, we randomly pick up an edge and walk through it to the next vertex. Meanwhile, with a certain possibility, the surf just jumps back to the start vertex. The possibility of reaching each vertex $v_i$ can be calculated as follows:

$$P(v_i) = \theta \cdot P(v_i = v_0) + (1-\theta) \cdot \sum_j \frac{a_{ji} \cdot P(v_j)}{\sum_k a_{jk}}$$

where $a_{ji} \in [0, 1]$ indicates whether there is a link between $v_j$ and $v_i$, and $\theta \in [0, 1]$ is the restart possibility to emphasize the vertices which are closely related with the start vertex $v_0$ ($\theta = 0.2$ in this study). The scores can be calculated recursively using the following formula:

$$\mathbf{v}^{n+1} = \theta \cdot \mathbf{v}^0 + (1-\theta) \cdot \mathbf{v}^n \mathbf{A}^*$$

where $\mathbf{v}^0$ is the initial vector, within which the element corresponding to the seed term is set to 1, and the others are all set to 0; $\mathbf{A}^*$ is the row-normalized adjacency matrix; and $\mathbf{v}^n$ is the weight vector of vertexes at the $n^{th}$ iteration. The iteration is repeated until $\mathbf{v}$ is converged, i.e. $|\mathbf{v}^{n+1} - \mathbf{v}^n| \le \sigma (\sigma = 0.001)$.

## 5. Experiment

*5.1 Experiment setup*

*5.1.1 Dataset*

Due to the novelty of problem definition, there is no public benchmark dataset yet. Thus we collect coordinate terms of fifty Chinese seed terms for evaluation[3]. Two annotators participate in the creation of gold answers. For each seed term, first the annotators manually emulate the possible concepts, and then for each concept they search the web using the query including the seed term and the concept name. The search results are manually checked and the coordinated terms are extracted from those pages. The results of the evaluated systems are also manually examined and the novel correct terms are added into the gold answers. The answers that both annotators agreed on are kept as the final dataset. Any result that is not contained in the gold answer is considered as not coordinate with the seed term. Because the Google Set system was shut down in 2011, it was only evaluated on only a part of our dataset[4]. We denote the small part of the dataset as Dataset 1, and the whole dataset as Dataset 2. The statistics of the gold results are shown in table 1.

Table 1 The statistics of the gold answers in the dataset

|  | Number of seed terms | Avg. concepts / seed term | Avg. coordinators / seed term |
|---|---|---|---|
| Dataset 1 | 15 | 1.6 | 154 |
| Dataset 2 | 50 | 1.3 | 118 |

*5.1.2 Evaluation metric*

In the experiment, we use two kinds of metrics to evaluate the results. One kind of metrics is term based, and the other kind is concept based. Note that the following metrics are calculated per seed term, and the mean values over all queries are reported in the evaluations.

*Term based metrics*

The term based metrics only take into account the extracted term itself, and ignore the underlying concept. A term is considered correct if it occurs in any list of different concepts in the gold answer. For example, 纽约/*New York* and 布什/*Bush* are both correct coordinate terms of 华盛顿/*Washington*. We use two metrics:

**Precision at n** (**P@n**) is the fraction of terms that are coordinate with the user's seed term in the top *n* results:

$$P@n = \frac{|\{\text{top } n \text{ extracted results}\} \cap \{\text{all gold results}\}|}{n}$$

The **Average Precision** (**AP**) is a widely used metric for ranking list evaluation. It considers the order in which the results are presented. It is the average of precision values computed at the point of each relevant result in the ranked sequence:

$$AveP(RL, GL) = \frac{1}{|GL|} \sum_{r=1}^{|RL|} cor(r) \cdot P@j$$

where *RL* is the result list for the query; |*RL*| is the number of terms in *RL*; *GL* is the gold answer list, |*GL*| is the number of terms in *GL*, *r* is the rank, and *cor*(*r*) is a binary function on the correctness of result at rank *r*.

These two metrics only perform on a single list of result. If the system returns several lists of term for a seed term, we merge them into a single list by take one term in each list in turn, and perform the evaluation on the merged list. E.g., if the results of system is $\{t_{11}, t_{12},…\}, \{t_{21}, t_{22}, …\}…\{t_{n1}, t_{n2}, ...\}$, then the merged list is $\{t_{11}, t_{21},… t_{n1}, t_{12}, t_{22},…t_{n2},…\}$.

*Concept based metrics*

The concept based metrics consider each result list as a concept. An extracted term is considered incorrect if it does not belong to the concept, no matter whether it is coordinate with the seed term. For example, 布什/*Bush* is an incorrect result in the list {布什/*Bush*, 纽约/*New York*, 芝加哥/*Chicago*, 洛杉矶/*Los Angles*}, because most terms in this list are cities.

---

[3] The dataset is available at http://www.icst.pku.edu.cn/lcwm/files/coordinate_terms_50.zip
[4] The dataset is also available at http://www.icst.pku.edu.cn/lcwm/files/coordinate_terms_15.zip

In the experiments, we calculate the weighted **Averages of AP** (**AAP**) over all clusters. Formally, let {$RL_1$, …, $RL_m$} be the result lists of the seed term, where each list represents a automatically mined concept, {$GL_1$,..., $GL_n$} be the gold answer lists, where each list represents a manually annotated concept, then AAP is computed by taking the weighted average of maximal average precision values:

$$AAP = \frac{1}{\sum_{i=1}^{m}|RL_i|}\sum_{i=1}^{m}|RL_i| \cdot \max_j AvgP(RL_i, GL_j)$$

AAP penalizes the noise in a cluster, but it does not reward grouping items from the same category together. **Inverse Average of AP** (**IAAP**) over all clusters focuses on the cluster with maximum average precision for each category. It is defined as:

$$IAAP = \frac{1}{\sum_{j=1}^{n}|GL_j|}\sum_{j=1}^{n}|GL_j| \cdot \max_i AvgP(RL_i, GL_j)$$

*5.1.3 Baseline systems*

In the experiment, we use two baseline systems:

**Google Sets** was a well-known and publicly accessible set expansion system in many languages. It is interesting to see how traditional set expansion system performs with only one seed term. The results were collected by manually submitting query terms to Google Sets before it was shut down.

**CoMiner** is a linguistic pattern based algorithm for mining competitors of a given entity (Bao et al. 2008; Li et al. 2006). The competitor names are coordinate with the given entity name. The original algorithm is designed for English, and we adapted the patterns as follows:

- H1: 例如 *EN*（、*CN*）* 或 ‖ 和 *CN*, e.g., "例如索尼、飞利浦和TDK" (*such as Sony, Philips and TDK*)
- H2: 特别是 *EN*（、*CN*）* 和（*CN*）, e.g., "特别是索尼、佳能和松下相机" (*especially Sony, Canon and Toshiba camera*)
- H3: 包括 *EN*（、*CN*）* 和（*CN*）, e.g. "知名品牌包括索尼、尼康和佳能" (*Leading brand including Sony, Nikon and Canon*)
- C1: *CN* 比 *EN* 更, e.g., "尼康比索尼更专业" (*Nikon is more professional than Sony.*)
- C2: *EN* 比 *CN* 更, e.g., "索尼比尼康更时尚" (*Sony is more fashion than Nikon.*)
- C3: *EN* 或 *CN*, e.g., "你可以选择索尼或佳能" (*You can choose Sony or Canon.*)
- C4: *CN* 或 *EN*, e.g., "佳能或索尼都可以" (*Either Canon or Sony is good choice.*)

In the patterns, *EN* denotes the user given seed entity, and *CN* denote a competitor name (i.e. a coordinate term). Note that *CN* must be at the end of a sentence in C3, otherwise its right boundary is not defined. For example,

求购索尼或佳能单反相机。(*Want a Sony or Canon Digital Singular Lens Reflex camera.*)

We cannot know where the end of the entity name is, i.e. the entity may be "佳", "佳能", "佳能单", etc. Similarly, *CN* must be at the beginning of a sentence in C4, otherwise its left boundary is not defined.

*5.2 Overall result*

Because Google Sets was closed in 2011, it is only evaluated in a small set of terms (i.e. Dataset 1). The evaluation results are shown in table 2. To evaluate whether the concept disambiguation is helpful, we report the results of our system with and without the concept disambiguation step (denoted as CTMS and CTMS-ND respectively). The evaluation results of CoMiner and our systems on the whole dataset are shown in table 3.

Overall, the evaluation results demonstrate that our proposed systems perform well in both precision and coverage. Both the systems with and without concept disambigation outperforms the two baseline systems over all metrics. In particular, the performance values of the system with concept disambuation (CTMS) are superior than the system without concept disambiguation (CTMS-ND). By adding the concept disambiguation step, the terms of nonprimary concepts are brought forward as new result lists which also have good quality. In the mean time, considering the concept-based metrics, the orginal list of terms is cleaned up by removing the terms of nonprimary concepts (which are considered incorrect in the orginal list), so its average precision increases.

The pattern based CoMiner method get a good P@10 value, but a low AP value. The main reason is that CoMiner only finds a small set of coordinate terms, and misses a lot of terms in the gold results. This problem also happens in GoogleSet. In additon, GoogleSet relies on the co-occurences among terms. However, the co-occurences can also be casued by many other reasons (e.g. modification relation, subject-object relation, etc.), and thus they are not reliable without validation. In fact, Google Sets even returns some meaningless string fragments, e.g, "发布: 5个月前" (*issued: 5 month ago*).

Table 2 The evaluation results of coordinate term mining systems on dataset 1

|  | P@10 | AP | AAP | IAAP |
|---|---|---|---|---|
| Google Sets | 0.393 | 0.106 | 0.285 | 0.229 |
| CoMiner | 0.621 | 0.140 | 0.250 | 0.210 |
| CTMS-ND | 0.840 | 0.605 | 0.738 | 0.629 |
| CTMS | **0.860** | **0.638** | **0.772** | **0.668** |

Table 3 The evaluation results of coordinate term mining systems on dataset 2

|  | P@10 | AP | AAP | IAAP |
|---|---|---|---|---|
| CoMiner | 0.674 | 0.127 | 0.231 | 0.193 |
| CTMS-ND | **0.900** | 0.571 | 0.622 | 0.596 |
| CTMS | 0.890 | **0.617** | **0.724** | **0.705** |

*5.3 Component analysis*

*5.3.1 Lingusitic clue word*

To analyze the contribution of each linguistic indicator function word, we run the linguistic based coordinate term extraction step with different clue word sets. The number of successfully retrieved coordinate terms (denoted as **Recall\***), the precision of all results (**Precision**) and the precision of top five results (**P@5**) are reported in table 4.

Among these clue words, 比 (*than*) leads to the most precise coordinators but the recall is not high enough. The precision values of other clue words are worse. However, the precision values of top five results are quite good, especially when we use {和 (*and*), 比 (*than*)} as clues. Thus this clue set is chosen in our system and the top five results are selected as initial coordinate terms.

Table 4 The performances of linguistic-based extraction using different clue words

| Clue Words | Recall* | Precision | P@5 |
|---|---|---|---|
| 和 (and) | 7.3 | 0.63 | 0.78 |
| 或 (or) | 4.6 | 0.73 | 0.66 |
| 比 (than) | 2.9 | 0.79 | 0.54 |
| 和+或 (and+or) | 9.8 | 0.55 | 0.81 |
| 和+比 (and+than) | 8.3 | 0.55 | 0.82 |
| 和+或+比 (and+or+than) | 10.6 | 0.50 | 0.82 |

*5.3.2 Extended seed terms for query*

The most important factor in the semi-structure based mining step is how many extended seed terms should be used in each query to find intended pages. Because the results of the linguistic based extractor can be noisy, or they may belong to different concepts, it will not help represent the concept by using too many terms in a query. If we do so, it will lead to fewer useful pages and more spam pages. The evaluation results in fig. 11 confirm our belief. Both metrics fall down when the number of query terms increases. For this reason, we use only two terms in each query, i.e. we choose each potential coordinate term plus the seed term to search for intended pages.

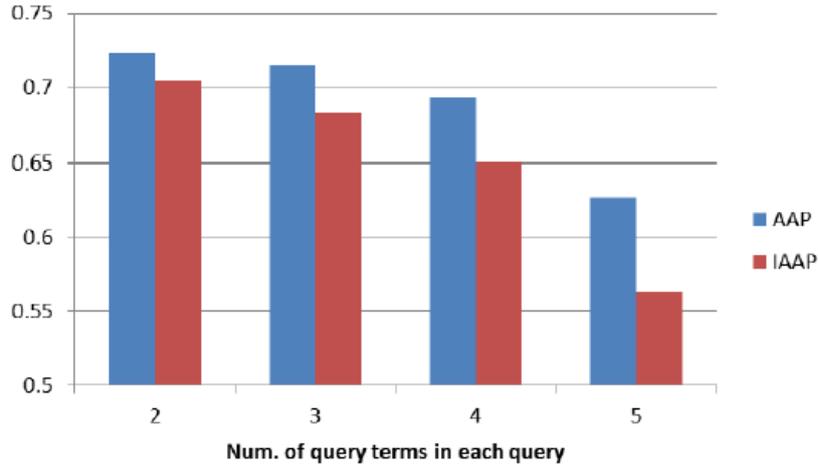

**Fig. 11**: Performances under different choices of query terms in semi-structure based extraction

*5.3.3 Concept disambiguation*

We use three metrics Purity, Inverse Purity, and F-measure (Manning et al. 2008) over clusters to evaluate the performance of the concept disambiguation step. All the seed terms in the dataset are used in this experiment, including those which has only one concept in the gold answer, because in practice the system is not able to know whether there are several concepts within the results in advance. Note that the terms which are not contained in the gold results are excluded in the evaluation. We also evaluate a naïve baseline model, i.e. grouping all terms into a single concept. The results are shown in table 4.

The performance values of all three systems are high. A potential reason is that the number of concepts is small in the dataset thus the task is not as hard as the traditional word sense disambiguation task. The system using content-based features only outperform the model using context-based features. The web pages usually contain some noisy information, such as navigation and advertisement, and thus the contexts of lists are less reliable. However, by integrating the content-based features and context-based features, the performance of concept disambiguation can be further improved.

*5.4 Example results*

Tables 5 ~ 9 illustrate some sample results of the systems. Due to space limitation, only top 10 results (excluding the seed term) are shown. Generally speaking, our system gives results in good accuracy and coverage. For the term 华盛顿 (*Washington*), our system finds a list of US Presidents and a list of cities. For the term 姚明 (*Yao Ming*), our system separates the NBA players from the Chinese athletes. Google Sets finds some coordinate terms for each seed, but it also finds other kinds of related terms. Sometimes it even returns meaningless text fragments, e.g. "打分: 2人" (*rating*: 2 *persons*) in the results of 宝马 (*BMW*). The results show that the single seed term does not provide enough information for the algorithm of Google Sets to generate an unambiguous term set. The top 10 results of the pattern based CoMiner are actually good, in spite of some small errors (e.g. 湖人/*Lakers* – 科比/*Kobe*). The major drawback of this method is that it can only extract incomplete results.

Table 5 The performances of concept disambiguation

|  | Purity | Inverse Purity | F-measure |
|---|---|---|---|
| Basline | 0.851 | 0.790 | 0.803 |
| Content-based | 0.935 | 0.829 | 0.844 |
| Context-based | 0.933 | 0.812 | 0.826 |
| Content + Context | 0.945 | 0.846 | 0.855 |

Table 6 The results of the seed term 华盛顿 (*Washington*)

| CEMS | | Google Sets | CoMiner |
|---|---|---|---|
| 林肯 (Lincoln) | 北京 (Beijing) | 纽约 (New York) | 费城 (Philadephia) |
| 杰斐逊 (Jefferson) | 洛杉矶 (Los Angeles) | 波士顿 (Boston) | 芝加哥 (Chicago) |
| 罗斯福 (Roosevelt) | 旧金山 (San Francisco) | 费城 (Philadelphia) | 莫斯科 (Moscow) |
| 杜鲁门 (Truman) | 芝加哥 (Chicago) | 新泽西 (New Jersey) | 洛杉矶 (Logs Angles) |
| 尼克松 (Nixon) | 格林斯波勒 (Greensboro) | 美中 (Middle of America) | 纽约 (New Yrok) |
| 里根 (Reagan) | 斯波坎 (Spokane) | 美南 (South of America) | 亚历山大 (Alexandria) |
| 肯尼迪 (Kennedy) | 亨茨维尔 (Huntsville) | 美西北 (Northwest of America) | 弗吉尼亚 (Virginia) |
| 布什 (Bush) | 尤金 (Eugene) | 澳洲 (Australia) | 西雅图 (Seattle) |
| 亚当斯 (Adams) | 弗雷斯诺 (Fresno) | 加拿大 (Canada) | 加州 (Califonia) |
| 艾森豪威尔 (Eisenhower) | 列克星敦 (Lexington) | 欧洲 (Europe) | 古巴 (Cuba) |

Table 7 The results of the seed term 姚明 (*Yao Ming*)

| CEMS | | Google Sets | CoMiner |
|---|---|---|---|
| 麦迪 (McGrady) | 刘翔 (Liu Xiang) | 火箭 (Rockets) | 刘翔 (Liu Xiang) |
| 奥尼尔 (O'Neal) | 王励勤 (Wang Liqing) | 麦迪 (McGrady) | 林书豪 (Lin Shuhao) |
| 纳什 (Nash) | 郭晶晶 (Guo Jinging) | 火箭队 (Rockets Team) | 易建联 (Yi Jianlian) |
| 安东尼 (Anthony) | 林丹 (Lin Dan) | 季后赛 (playoffs) | 詹姆斯 (James) |
| 艾佛森 (Iverson) | 易建联 (Yi Jianlian) | 会员 (member) | 马布里 (Marbury) |
| 海耶斯 (Hayes) | 邱凯 (Qiu Kai) | 爵士 (Jazz) | 王楠 (Wang Nan) |
| 吉诺比利 (Ginobili) | 杨威 (Yang Wei) | 评论 (comment) | 马丁 (Martin) |
| 朗多 (Rondo) | 马琳 (Ma Lin) | 科比 (Kobe) | 麦迪 (McGreedy) |
| 雷阿伦 (Ray Allen) | 张怡宁 (Zhang Yining) | NBA | 布鲁克斯 (Brooks) |
| 詹姆斯 (James) | 运动员 (athlete) | 发布: 7个月前 (released: 7 months ago) | 菲尔普斯 (Phelps) |

**Table 8** The results of the seed term 湖人 (*Lakers*)

| CEMS | Google Sets | CoMiner |
|---|---|---|
| 火箭 (Rockets) | 火箭 (Rockets) | 快船 (Clippers) |
| 凯尔特人 (Celtics) | 姚明 (Yao Ming) | 热火 (Heats) |
| 热火 (Heats) | 科比 (Kobe) | 魔术 (Magics) |
| 魔术 (Magics) | 麦迪 (McGrady) | 科比 (Koby) |
| 马刺 (Spurs) | 季后赛 (playoffs) | 雷霆 (Thuder) |
| 活塞 (Pistons) | 爵士 (Jazz) | 斯洛文尼亚 (Slovenia) |
| 小牛 (Mavericks) | 火箭队 (Rockets Team) | 保罗 (Paul) |
| 爵士 (Jazz) | 太阳 (Suns) | 爵士 (Jazz) |
| 太阳 (Suns) | 篮球 (basketball) | 小牛 (Mavericks) |
| 雷霆 (Thunder) | 帖子中包含的关键字 (Keywords in post) | 公牛 (Bulls) |

**Table 9** The results of the seed term 宝马 (*BMW*)

| CEMS | Google Sets | CoMiner |
|---|---|---|
| 奥迪 (Audi) | 帖子中包含的关键词 (keywords in post) | 奔驰 (Benz) |
| 奔驰 (Benz) | 奔驰 (Benz) | 沃尔沃 (Volve) |
| 本田 (Honda) | 发布: 5个月前 (issued: 5 month ago) | 霸道 (Prado) |
| 丰田 (Toyota) | 打分: 2人 (rating: 2 persons) | 华晨 (Brilliance) |
| 大众 (VW) | 会员 (member) | 奥迪 (Audi) |
| 保时捷 (Porsche) | 车 (vehicle) | 丰田 (Toyota) |
| 福特 (Ford) | 时长 (time-length) | 本田 (Honda) |
| 马自达 (Mazda) | 视频 (video) | 迈腾 (Magotan) |
| 别克 (Buick) | 奥迪 (Audi) | 萨博 (Saab) |
| 标致 (Peugeot) | 奔驰车 (Benz car) | 路虎 (Land Rover) |

**6. Application on named entity recognition**

In this section, we briefly introduce an application of our proposal on Named Entity Recognition (NER) in Chinese News Comments. For full details refer to (Wan et al. 2011).

The task aims to extract named entities (i.e. Peoples, Locations, and Organizations) from user generated comments for a news article. Because news comments are freely written by different persons with different education background and writing styles, they are very different from formal news text. In particular, named entities in news comments are usually composed of some wrongly written words, informal abbreviations or aliases, which brings great difficulties for machine detection and understanding. Fortunately, most news comments are relevant to the news topics in the referred news article, and thus most entities in the news comments are related to the entities in the news article. Considering these close relationships, the entity information in the news article can be used to improve named entity recognition in the news comments.

Fig. 12 shows the framework of the proposed NER approach. The basic idea is to find a few useful named entities in or related to the news article, and then incorporate the entity information into the basic CRF-based NER tagging algorithm. Three kinds of useful entities are investigated, including

- All NE, including all the named entities in the news article.
- Focused NE, i.e. the named entities which are most relevant to the main topic of the news article.
- Related NE, i.e. the coordinate entities of the focused entities.

The related named entities are useful because the news comments usually refer to some related or alternated entities of the focused entities. For example, when a news article is talking about "中国移动/*China Mobile*", the associated comments may mention "中国联通/China Unicom" or "中国电信/China Telecom". A common kind of related NE is the coordinate entity, which can be extracted by our proposed approach, using each focused NE as the seed term. The experimental results in (Wan et al. 2011) have shown the effectiveness of using coordinate named entities for NER in Chinese News comments.

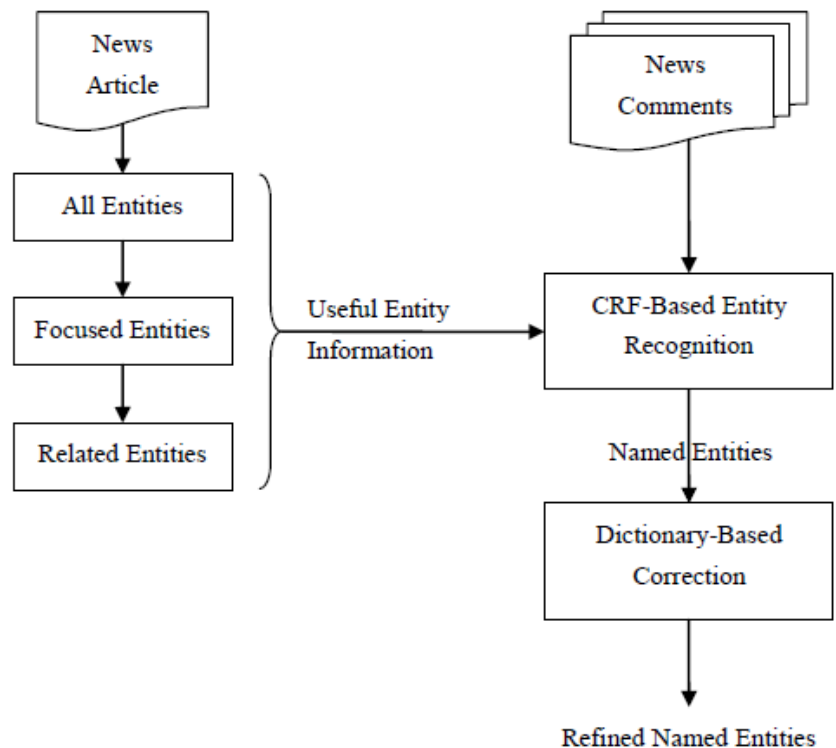

**Fig. 12**: The framework of named entity recognition system in Chinese news comments

**7. Conclusion and future work**

In this paper, we focus on the task of mining coordinate terms of a user given Chinese seed term. In our approach, we integrate manually defined linguistic patterns and automatically learned semi-structural templates together to extract coordinate terms. In addition, we group the terms into different concepts and rank the terms according to their significance. The experimental results show that our system generates results in both high quality and high coverage. The linguistic based extractor performs best when 比 (*than*) and 和 (*and*) are used as clue words. The semi-structure based extractor performs best when the user-given seed term together with each of top five results of linguistic based extractor are used as query to the search engine. The concept disambiguation step can further improve the performance values.

In future, we will utilize machine learning algorithms to find more linguistic patterns. We also plan to extract the concept names of term clusters from the context and online resources. The further studies of the coordinate relation, e.g. mining the commonalities and differences between those terms, are also interesting research tasks.